\documentclass[pdflatex,sn-mathphys-num]{sn-jnl}

\usepackage{multirow}%
\usepackage{amsmath,amssymb,amsfonts}%
\usepackage{amsthm}%
\usepackage{mathrsfs}%
\usepackage[title]{appendix}%
\usepackage{xcolor}%
\usepackage{textcomp}%
\usepackage{manyfoot}%
\usepackage{booktabs}%
\usepackage{algorithm}%
\usepackage{algorithmicx}%
\usepackage{algpseudocode}%
\usepackage{listings}%
\usepackage{graphicx}%
\usepackage{hyperref}
\usepackage{lmodern}
\usepackage{makecell}
\usepackage{anyfontsize}

\begin{document}

\title[Article Title]{Feature Complementation Architecture for Visual Place Recognition}


\author[1]{\fnm{Weiwei} \sur{Wang}}\email{231611014@sust.edu.cn}
\equalcont{These authors contributed equally to this work.}

\author[1]{\fnm{Meijia} \sur{Wang}}\email{4672@sust.edu.cn}
\equalcont{These authors contributed equally to this work.}

\author[1]{\fnm{Haoyi} \sur{Wang}}\email{haoyi18700580653@163.com}

\author[1]{\fnm{Wenqiang} \sur{Guo}}\email{guowenqiang@sust.edu.cn}

\author[2]{\fnm{Jiapan} \sur{Guo}}\email{j.guo@rug.nl}

\author[3]{\fnm{Changming} \sur{Sun}}\email{changming.sun@csiro.au}

\author*[1]{\fnm{Lingkun} \sur{Ma}}\email{malingkun@sust.edu.cn}

\author*[1]{\fnm{Weichuan} \sur{Zhang}}\email{zwc2003@163.com}

\affil*[1]{\orgdiv{School of Electronic Information and Artificial Intelligence}, \orgname{Shaanxi University of Science and Technology}, \city{Xi'an}, \postcode{710021}, \state{Shaanxi Province}, \country{China}}

\affil[2]{\orgdiv{Department of Radiation Oncology}, \orgname{University Medical Center Groningen, University of Groningen}, \orgaddress{\street{Hanzeplein 1}, \city{Groningen}, \postcode{9713GZ}, \country{The Netherlands}}}

\affil[3]{\orgdiv{CSIRO Data61}, \orgaddress{\street{PO Box 76}, \city{Epping}, \state{NSW},  \postcode{1710}, \country{Australia}}}


\abstract{Visual place recognition (VPR) plays a crucial role in robotic localization and navigation. The key challenge lies in constructing feature representations that are robust to environmental changes. Existing methods typically adopt convolutional neural networks (CNNs) or vision Transformers (ViTs) as feature extractors. However, these architectures excel in different aspects—CNNs are effective at capturing local details. At the same time, ViTs are better suited for modeling global context, making it difficult to leverage the strengths of both. To address this issue, we propose a local-global feature complementation network (LGCN) for VPR which integrates a parallel CNN-ViT hybrid architecture with a dynamic feature fusion module (DFM). The DFM performs dynamic feature fusion through joint modeling of spatial and channel-wise dependencies. Furthermore, to enhance the expressiveness and adaptability of the ViT branch for VPR tasks, we introduce lightweight frequency-spatial fusion adapters into the frozen ViT backbone. These adapters enable task-specific adaptation with controlled parameter overhead. Extensive experiments on multiple VPR benchmark datasets demonstrate that the proposed LGCN consistently outperforms existing approaches in terms of localization accuracy and robustness, validating its effectiveness and generalizability.}




\maketitle

\section{Introduction}\label{sec1}
Visual place recognition (VPR), also known as visual geolocalization \cite{1,11,tang2025cascading}, aims to retrieve the geographical location of a query image captured by a visual sensor from a known geo-referenced database. As a key component in simultaneous localization and mapping (SLAM) and global relocalization, VPR plays a central role in various applications such as city-scale navigation for autonomous vehicles \cite{2,10,liao2025dynamic}, autonomous path planning for drones \cite{jing2022recent,zhang2020corner}, and scene alignment in augmented reality systems \cite{4}. However, achieving robust VPR in open and complex environments remains a significant challenge, primarily due to two factors: appearance variation, caused by changes in lighting, weather, and seasons; and viewpoint change, resulting from differences in camera angles or trajectory shifts. These factors can severely degrade the stability and robustness of feature matching.

\begin{figure}[htbp]
\centering
\includegraphics[width=\textwidth,keepaspectratio]{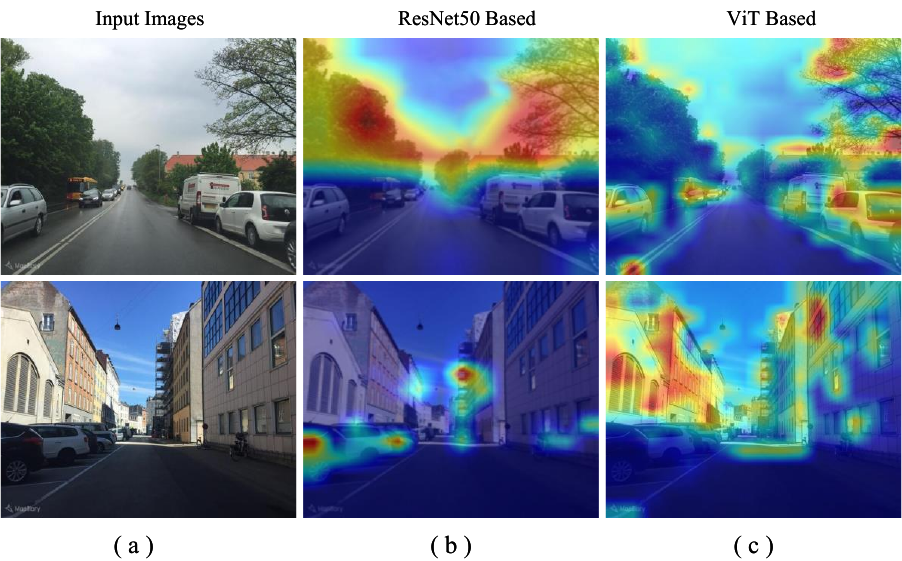}
\caption{Heatmap visualization of feature representations obtained by different backbone networks for distinct input images: (a) Two input images; (b) Attention region visualization by ResNet50; (c) Attention region visualization by ViT.}\label{fig1}
\end{figure}

Recent advances in deep learning-based feature representations have substantially improved VPR performance \cite{5,6,7,8,9,47,48,49,wang2025principal,zhang2019discrete, wang2024unbiased,lei2024semi,lu2023track,zhang2017noise,zhang2014corner,jing2022image, shui2013corner,zhang2015contour,zhang2021ndpnet,ren2024few}. Convolutional neural networks (CNNs), with their local receptive fields and translation invariance, are effective at capturing fine-grained structures and local discriminative features such as textures and edges. In contrast, vision Transformers (ViTs) leverage self-attention mechanisms to model long-range dependencies and global context, demonstrating strong adaptability across diverse environments. However, each architecture has limitations, and it is difficult for a single model to fully capture both local details and global semantics. As illustrated in the heatmap examples in Fig. \ref{fig1}, this inherent difference in attention patterns can lead to structural biases, resulting in mismatches and performance degradation in complex scenes.
To address these issues, we propose a local-global feature complementation network (LGCN), which combines the complementary strengths of CNNs and ViTs in modeling local details and global semantics. Specifically, we design a parallel ViT-CNN hybrid architecture and introduce a dynamic feature fusion module (DFM). The DFM employs cascaded 1×1 convolutions and nonlinear activation functions to generate spatial-channel attention weights, which are then modulated by a learnable global scaling factor. This enables dynamic control over the fusion ratio between the two streams, enhancing discriminative regions while suppressing irrelevant noise to improve overall robustness.

In addition to improving the adaptability of the frozen ViT backbone to VPR tasks, we introduce frequency-spatial fusion adapter modules into each Transformer block. These lightweight adapters are inserted after the attention layer and operate in parallel with the feedforward network. The module incorporates both frequency and spatial branches to introduce local inductive bias while maintaining global modeling capability. The frequency branch uses the Fourier transform to modulate amplitude features, enhancing structural information, while the spatial branch applies depthwise separable convolutions to emphasize local details. The outputs of the two branches are fused into a residual term, which is then added to the main ViT output, enabling dynamic feature enhancement without retraining the backbone. This design significantly improves the model’s discriminative power and generalization in challenging environments.

In summary, our main contributions are as follows:
\begin{itemize}
    \item Local-global feature complementation network: We propose a parallel ViT-CNN framework that integrates features from different backbones using spatial resampling and channel alignment for effective multi-scale fusion.
    \item Dynamic feature fusion mechanism: We design a lightweight DFM module that dynamically adjusts fusion weights in spatial and channel dimensions via pixel-level attention and a global scaling factor, thereby enhancing the response to discriminative regions.
    \item Frequency-spatial fusion adapter: We introduce adapter modules into the frozen ViT backbone to compensate for its limited local inductive bias, using complementary frequency and spatial modeling to improve adaptability.
    \item Extensive empirical validation: We evaluate our method on several cross-domain VPR benchmarks, including Pitts30k, Nordland, and MSLS. The results show that our approach outperforms existing state-of-the-art methods, demonstrating superior robustness and generalization.
\end{itemize}

The remainder of this paper is organized as follows: Section. \ref{sec2} reviews related work. Section. \ref{sec3} presents the LGCN and its core components. Section. \ref{sec4} details experimental evaluations and results. Section. \ref{sec5} concludes the paper and discusses future research directions.
\begin{figure}[htbp]
\centering
\includegraphics[scale=0.9]{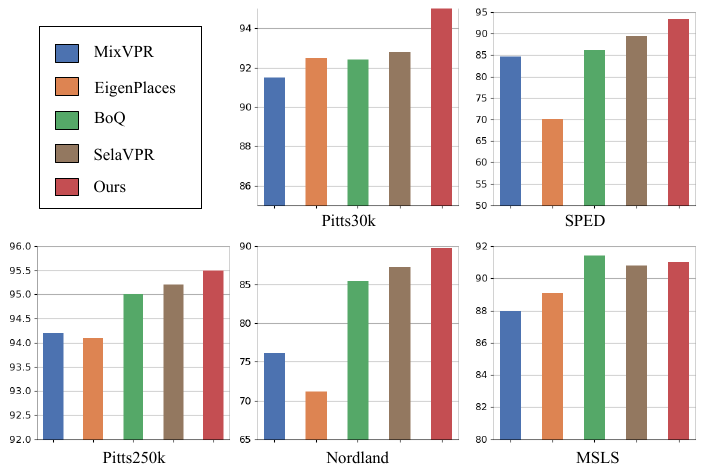}
\caption{Comparison of our method with the state-of-the-art methods MixVPR, EigenPlaces, BoQ, and SelaVPR in terms of Recall@1 performance.}\label{fig2}
\end{figure}
\section{Related Work}
\label{sec2}
\subsection{VPR Methods Based on CNN}
\label{subsec2_1}
The feature extraction architectures for VPR have evolved from handcrafted designs to deep learning-based approaches. Early studies primarily relied on handcrafted feature extraction methods such as SIFT \cite{17}, SURF \cite{18}, and ORB \cite{19}, which exhibited limitations in handling viewpoint and illumination variations. With the rise of deep learning, convolutional neural networks (CNNs) were introduced into the VPR domain, significantly improving recognition accuracy through their powerful feature extraction capabilities. 

The representative NetVLAD \cite{12} enhanced VPR performance by introducing a trainable VLAD layer to aggregate CNN features into compact global descriptors. Numerous NetVLAD variants have since emerged. For instance, Patch-NetVLAD \cite{20} significantly improved robustness to viewpoint and environmental changes by extracting multi-scale local features from the residuals of the global NetVLAD descriptor and leveraging them for matching and fusion. SSR-VLAD \cite{21} achieved real-time VPR without GPU acceleration through a semantic skeleton representation (SSR) and spatiotemporal aggregation framework. CosPlace \cite{23} transformed the VPR task into a classification problem, avoiding computationally expensive negative sample mining in contrastive learning, thereby enhancing training efficiency and addressing scalability challenges in large-scale scenarios. MixVPR \cite{24} generated robust global descriptors by iteratively integrating global feature relationships through a fully-connected layer stacking structure, outperforming existing methods with fewer parameters and faster inference. EigenPlaces \cite{25} improved VPR performance by clustering training data into categories containing multi-view images of the same scene, enabling the training of viewpoint-robust global descriptors while reducing GPU memory requirements and descriptor dimensionality.

However, these methods remain constrained by the local receptive fields of CNNs, struggling to model cross-region global semantic relationships (e.g., ambiguous matches in repetitive structural scenes).
\subsection{VPR Methods Based on ViT}
\label{subsec2_2}
The Transformer architecture, initially groundbreaking in natural language processing \cite{26}, was successfully adapted to computer vision tasks, giving rise to vision Transformers (ViTs) \cite{7}. In VPR, ViT is commonly integrated with generalized mean (GeM) pooling  \cite{22} and NetVLAD \cite{12} as a backbone network, leveraging self-attention mechanisms to model global contextual dependencies and demonstrating superior cross-domain generalization compared to CNNs.

The representative R$^2$Former \cite{27} validated ViT’s strong generalization in complex scenes through a dual-stage retrieval framework that utilizes feature correlations, attention weights, and spatial coordinates extracted by ViT for image pair similarity measurement. To enhance ViT’s fine-grained feature modeling, subsequent studies proposed multi-path improvements: SelaVPR \cite{15} suppressed dynamic object interference (e.g., pedestrians/vehicles) via a global-local feature calibration mechanism; CricaVPR \cite{28} introduced a cross-image correlation-aware module to align multi-view features within batches using attention maps, improving robustness to seasonal/illumination variations; BoQ \cite{28} designed learnable global query vectors to selectively aggregate discriminative regional features through cross-attention. Additionally, AnyLoc \cite{29} achieved efficient open-scene localization by combining large-scale pre-trained ViT's dense pixel-level features with unsupervised aggregation strategies, while ProGEO \cite{31} enhanced geo-semantic representation by fusing CLIP’s multimodal priors and dynamically optimizing the image encoder via text prompts.

However, ViT’s patch-wise processing may disrupt local texture continuity (e.g., fractured brick wall patterns) and exhibit weak feature responses in low-texture regions (e.g., skies/water surfaces), limiting its localization accuracy in texture-deficient scenarios.
\subsection{VPR Methods Based on Hybrid Architectures}
\label{subsec2_3}
Single-architecture approaches inherently lack complementary feature representations. Existing hybrid methods partially validate the complementary potential of heterogeneous architectures. For example, ETR \cite{32} employs a pre-trained CNN to extract global and local descriptors, models intra-image contextual relationships of local features via self-attention, and constructs cross-image similarity metrics using cross-attention, enabling retrieval and re-ranking in a single forward pass. HybridVPR \cite{16} fuses dual-stream CNN-Transformer features, aggregates multi-scale information using a semantic-aware NetVLAD layer, and enhances robustness to appearance and viewpoint variations with adaptive triplet loss. TransVPR \cite{14} generates global descriptors through multi-level attention fusion of CNN-extracted local primitive features and outputs key local descriptors to support geometric verification.

Despite these advancements, their fusion mechanisms still rely on static weight allocation strategies (e.g., fixed-ratio summation or channel concatenation), failing to dynamically adjust feature contribution weights based on scene characteristics (e.g., enhancing CNN responses in low-texture regions or reinforcing ViT’s discriminative power in repetitive structural areas). This leads to suboptimal representations in dynamic interference or complex urban scenarios, constraining the model’s generalization capability and robustness.
\subsection{Application of Parameter-Efficient Fine-Tuning in VPR}
\label{subsec2_4}
In VPR tasks, pre-trained visual models typically contain a large number of parameters. Direct fine-tuning often leads to overfitting and incurs significant computational costs. Recently, parameter-efficient fine-tuning (PEFT) methods have emerged as a promising approach to adapt large models to downstream tasks with minimal parameter updates \cite{41,42}. However, the application of PEFT in the field of VPR is still in its early stages. Some recent works have begun to incorporate adapter mechanisms into VPR models. For example, SelaVPR \cite{15} and CricaVPR \cite{28} integrate lightweight visual adapter modules into their Transformer backbones to enhance feature modeling and improve generalization. Nonetheless, these approaches mainly focus on spatial-domain adaptation and have not fully explored the potential of frequency-domain features within adapters.

Considering that VPR heavily relies on structural characteristics and repetitive texture patterns of scenes, solely modeling spatial-domain features may be insufficient for achieving stable and robust image matching under challenging conditions such as drastic environmental changes, weak textures, or repetitive structures. In contrast, frequency-domain information offers stronger global structural awareness and can effectively capture scale variations and repeated patterns.

Therefore, designing adapter modules that fuse frequency and spatial domain information is promising for improving the scene adaptability and localization accuracy of ViT-based backbones in VPR tasks. Developing parameter-efficient transfer mechanisms with frequency-domain awareness represents a future direction toward making VPR models more lightweight and robust.
\section{Methodology}
\label{sec3}
\subsection{Local-global Feature Complementation Network}
\label{subsec3_1}
VPR tasks are challenged by issues such as viewpoint changes, texture repetition, and textureless regions. These factors impose stringent requirements on feature extraction networks, which must be capable of capturing both local structural details and global semantic context. However, networks based on a single architecture often struggle to meet both demands. CNNs are well-suited for modeling fine-grained local features due to their local receptive fields and translation invariance, but their limited receptive scope hampers the modeling of long-range dependencies. In contrast, ViTs leverage self-attention to model global dependencies across the image, showing stronger cross-domain generalization. Nonetheless, the patch-based partitioning in ViT can disrupt local continuity, making it less effective in encoding fine structures or textureless areas.

\begin{figure}[htbp]
\centering
\includegraphics[width=\textwidth,keepaspectratio]{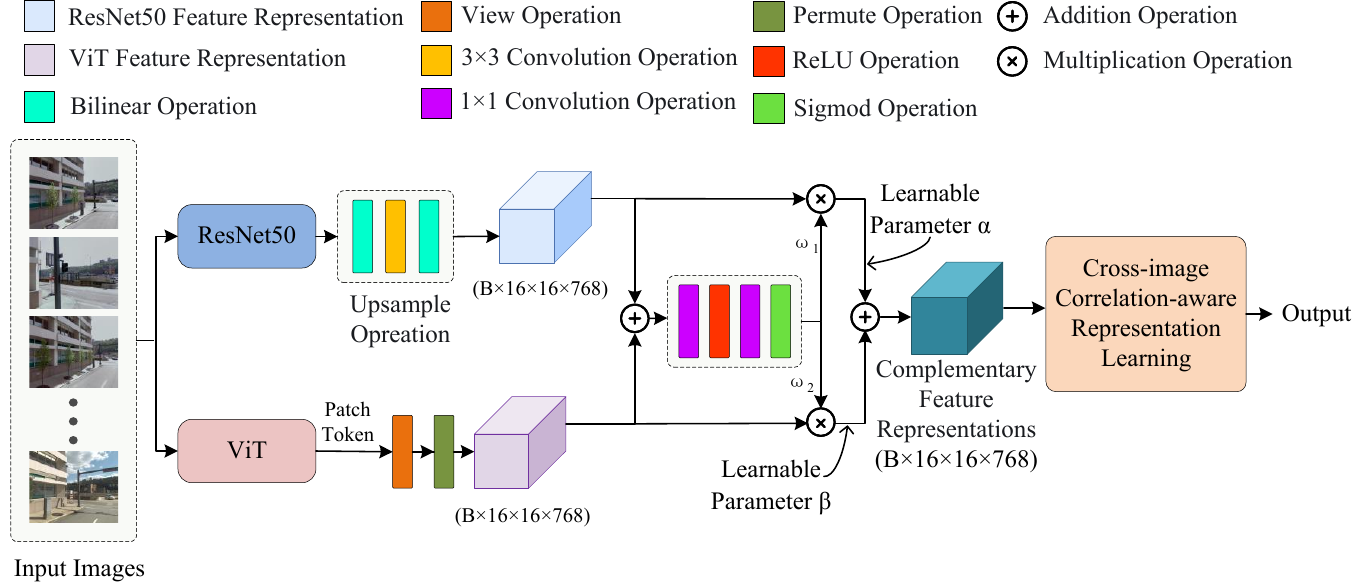}
\caption{Local-global feature complementation network framework.}\label{fig3}
\end{figure}

To leverage the complementary strengths of CNN and ViT, we propose a local-global feature complementation network. By constructing and integrating CNN and ViT feature paths in parallel, the model forms a unified heterogeneous feature space at the same resolution, enhancing the ability to represent complex scenes. The overall architecture is illustrated in Fig. \ref{fig3}.

\textbf{ViT Branch:} We adopt the DINOv2-pretrained ViT-Base model as the backbone for global feature extraction. The input image is divided into fixed-size patches of 16×16. After linear projection, patch tokens $F_{\mathrm{ViT}}\in R^{16\times16\times768}$ are generated. multi-head self-attention (MHSA) is then used to model long-range spatial dependencies. The attention mechanism within MHSA follows the standard formulation:
\begin{equation} 
\mathrm{Attention}(Q,K,V)=\mathrm{softmax}\bigg(\frac{QK^T}{\sqrt{d_k}}\bigg)V.
\end{equation}
\textbf{ResNet Branch:} We employ the DINO-pretrained ResNet-50 as the local feature extractor. Specifically, we retain the output of Stage 4 and discard Stage 5 to reduce model complexity and enhance mid-level semantic representations. The feature map output from this branch has a spatial size of 7$\times$7$\times$1024. To align with the ViT output in both spatial and channel dimensions, we design an upsampling module $\mathrm{U}(\cdot)$.The transformation is expressed as:
\begin{equation}
F'_{\text{Res}} = U(F_{\text{Res}}) = \text{Bilinear}_{14\times14} \rightarrow \text{Conv}_{3\times3}^{1024 \rightarrow 768} \rightarrow \text{Bilinear}_{16\times16}.
\end{equation}
The resulting feature map $F_{\mathrm{Res}}^{\prime}\in R^{16\times16\times768}$ is fully aligned with the ViT output in terms of spatial resolution and channel dimensions, providing a consistent representation basis for the subsequent cross-branch fusion.
\begin{figure}[htbp]
\centering
\includegraphics[width=\textwidth,keepaspectratio]{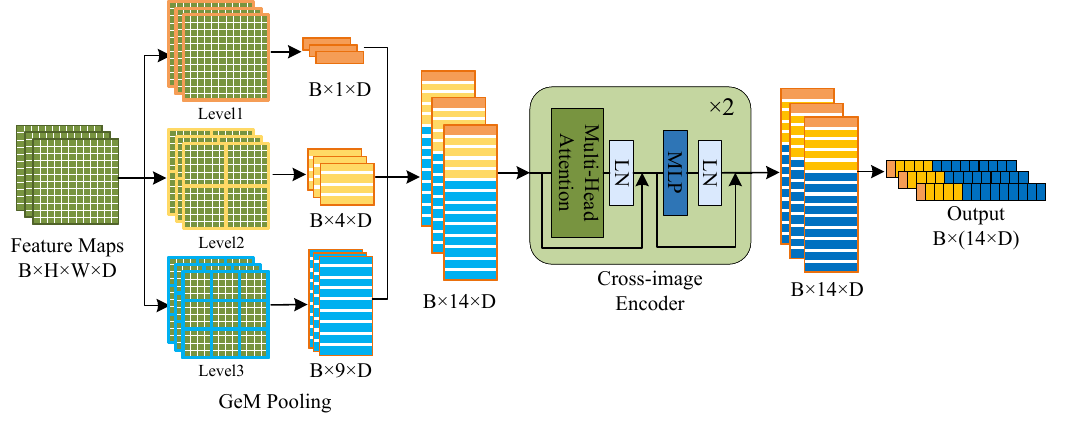}
\caption{Cross-image correlation-aware place representation moudle. }\label{fig4}
\end{figure}
\subsection{Dynamic Feature Fusion Module}
\label{subsec3_2}
Conventional feature fusion methods, such as channel concatenation or weighted averaging, typically rely on static fusion strategies. These approaches fail to adaptively adjust the fusion ratio in response to dynamic scene attributes (e.g., illumination, viewpoint, and occlusion), thereby limiting the complementary representation capabilities of heterogeneous features such as those from CNN and ViT in open-world environments. To address this limitation, we propose a dynamic feature fusion module. Based on channel attention mechanisms \cite{33}, this lightweight gated network models attention across both spatial and channel dimensions, enabling dynamic and selective feature fusion.

\textbf{Feature Fusion and Gated Attention Modeling.} We begin by performing element-wise addition of the feature maps from the CNN and ViT branches to obtain an initial fused representation:
\begin{equation}
F=F_{\mathrm{ViT}}+F_{\mathrm{Res}}^{\prime}.
\end{equation}
This operation integrates both local and global semantic cues, serving as input to the subsequent attention generation network.

Next, we design a bottleneck-style gating sub-network to model the importance of different channels. It consists of two 1$\times$1 convolutional layers for channel compression and excitation:
\begin{equation}
    \omega=\sigma(w_2\cdot\delta(w_1\cdot F)).
\end{equation}
Here, $\omega_{1}\in R^{768\times192}$ performs channel compression (reducing dimensionality to 1/4 to reduce computation), $\delta(\cdot)$ denotes the ReLU activation introducing non-linearity, and $\omega_{2}\in R^{192\times768}$ restores the original channel dimension. The sigmoid function $\sigma(\cdot)$ constrains the attention weights $\omega\in R^{16\times16\times768}$ to the [0, 1] range.

\textbf{Dynamic Feature Recombination.} The generated attention weights are used to guide the recombination of the fused features as follows:
\begin{equation}\begin{aligned}
    &F_{\mathrm{fused1}}=\omega\odot F_{\mathrm{ViT}},\\
     &F_{\mathrm{fused2}}=(1-\omega)\odot F_{\mathrm{ViT}},   
\end{aligned}\end{equation}
where $\odot$ denotes element-wise multiplication. This mechanism allows the network to emphasize ViT features in areas such as building contours and structural edges, where semantic context is important, while enhancing the contribution of CNN features in regions rich in texture or affected by occlusion, thus achieving more precise and robust feature selection.

To further improve fusion flexibility, we introduce two learnable parameters $\alpha_1$ and $\alpha_2$, which are optimized via backpropagation. These parameters allow the model to automatically balance the contributions of CNN and ViT branches during training, eliminating the need for hand-crafted fusion rules and enhancing the network’s adaptability across diverse environments.
\subsection{Frequency-Spatial Adapter Module}
\label{subsec3_3}
\begin{figure}[htbp]
\centering
\includegraphics[scale=1,keepaspectratio]{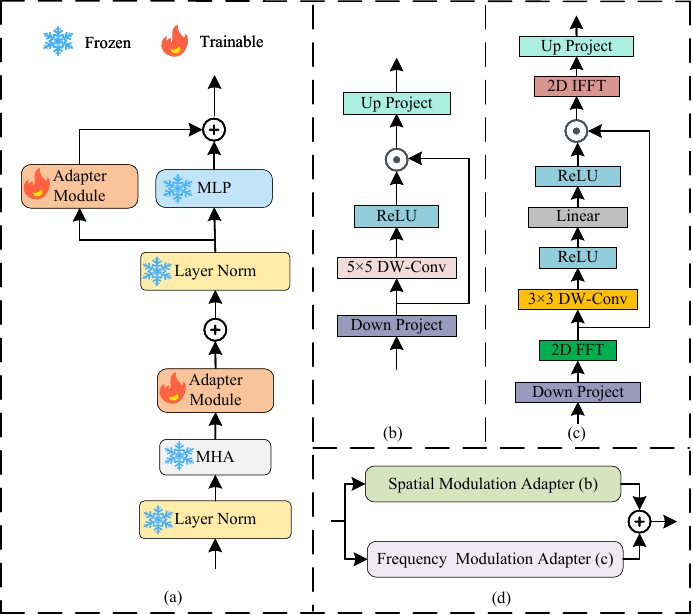}
\caption{Architecture of the frequency-spatial adapter (FSA) module. (a) Transformer block, (b) Spatial modulation adapter, (c) Frequency modulation adapter, (d) Adapter module.}\label{fig5}
\end{figure}

Images inherently contain rich frequency information. High-frequency components typically correspond to edges and textures, capturing fine-grained details, while low-frequency components represent global shapes and structural contours. In visual place recognition tasks, frequency-domain features play a crucial role in distinguishing subtle structures such as building facades and road patterns. Inspired by approaches in \cite{44,45,46}, we introduce a frequency-spatial adapter (FSA) module, which incorporates a frequency-aware pathway to enhance the model's robustness in complex environments while maintaining parameter efficiency. As illustrated in Fig. \ref{fig5}, the proposed adapter augments the backbone with additional frequency-domain modeling capabilities, enabling more comprehensive feature representation.

Specifically, we first extract patch tokens from the output of the ViT backbone and reshape them into a two-dimensional feature map to facilitate convolutional operations. A dual-path modulator is then applied to enhance the features in both spatial and frequency domains. In the spatial branch, depthwise convolutions are used to capture local contextual dependencies, emphasizing edge contours and texture details. In the frequency branch, the feature map is transformed into the frequency domain, where modulation is performed on the amplitude spectrum to improve robustness against global variations such as lighting changes and texture noise. The enhanced representation is then transformed back to the spatial domain via the inverse Fourier transform. Finally, the outputs from the two branches are fused along the feature dimension and added to the original features through a residual connection. This design enables dynamic feature reconstruction and fine-grained adaptation, contributing to improved expressiveness and robustness in diverse environments.
\section{Experiments}
\label{sec4}
\subsection{Datasets}
\label{subsec4_1}
We conduct experiments on seven commonly used VPR datasets, including Pitts250k, Pitts30k, Mapillary Street-Level Sequences (MSLS), SPED, Nordland, Eynsham, and St. Lucia. Our evaluations are based on established VPR benchmarks \cite{34}, and the key characteristics of each dataset are summarized in Table \ref{tab1}.
\begin{table}[htbp]
\centering
\caption{Summary of the datasets in experiments. \checkmark indicates the presence of this challenge.}
\begin{tabular}{cccccc}
\toprule
\multirow{2}{*}{Dataset} & \multicolumn{2}{c}{Number} & \multicolumn{3}{c}{Challenge}     \\ 
\cmidrule(lr){2-3} \cmidrule(lr){4-6}
                         & Database     & Queries     & Viewpoint & Season & Illumination \\ \hline
Pitts250k                & 8.2k         & 84k         & \checkmark         &        &              \\
Pitts30k                 & 10k          & 6,816       & \checkmark         &        &              \\
MSLS                     & 19k          & 740         & \checkmark         & \checkmark      & \checkmark            \\
SPED                     & 607          & 607         &           & \checkmark      & \checkmark            \\
Nordland                 & 27.6k        & 2,760       &           & \checkmark      &              \\
Eynsham                  & 24k          & 24k         & \checkmark         &        &              \\
St-Lucia                 & 1,464        & 1,549       &           &        &              \\ 
\bottomrule
\end{tabular}
\label{tab1}
\end{table}
Specifically, the Pitts250k \cite{35} and Pitts30k datasets were collected in the dense urban area of Pittsburgh using multi-trajectory vehicle-mounted cameras under varying conditions such as day/night and different weather (clear, rain, and snow). These datasets are primarily designed to evaluate feature stability under extreme viewpoint changes and dynamic occlusions. The SPED dataset \cite{36} comprises low-resolution images with deep scene layouts captured from surveillance cameras worldwide. It includes diverse variations in lighting, weather, and seasonal conditions, posing challenges in long-term place recognition. The MSLS\_val dataset \cite{37} covers time-sequenced street-level imagery from 11 cities across different continents. It reflects architectural diversity and long-term appearance changes (e.g., vegetation cycles spanning over 3 years), and is used to assess cross-city generalization performance. The Nordland dataset \cite{38} consists of seasonal imagery captured along the same railway route in Northern Europe. It tests structural consistency by comparing summer database images against winter query images, where dramatic changes such as vegetation loss and snow coverage are present. The Eynsham dataset \cite{39} focuses on suburban roads in the UK, including open farmland and large-scale repetitive vegetation. It is designed to evaluate discriminability under low-texture, homogeneous environments. The St. Lucia dataset \cite{40} captures subtropical suburban streets across different times during daytime. It emphasizes temporal variations such as gradual lighting transitions and dynamic tree shadows, offering a benchmark for evaluating robustness to time-varying conditions.
\subsection{Implementation Details}
\label{subsec4_2}
Our experiments are implemented based on a dual-stream heterogeneous architecture, utilizing pre-trained ViT-B/14 and ResNet-50 backbones for feature extraction. All experiments are conducted using PyTorch on an NVIDIA GeForce RTX 3090 GPU. The Adam optimizer is employed with a learning rate of 0.00001 and a batch size of 16. Inspired by CricaVPR \cite{28}, we further feed the features fused by DFM into a cross-image correlation-aware module to generate the final global image descriptor (as illustrated in Fig. \ref{fig4}). In our setup, positive samples are defined as reference images located within 10 meters of the query image, while negative samples are those located more than 25 meters away. The model is fine-tuned on the GSV\_Cities dataset \cite{43} and evaluated on benchmarks such as Pitts30k\_test. We assess retrieval performance using Recall@N (R@N), where metrics such as R@1, R@5, and R@10 indicate the probability that a correct match is found within the top-N retrieved candidates.
\subsection{Comparison with Previous Works}
\label{subsec4_3}
In this section, we conduct a comprehensive comparison between our proposed LGCN method and a range of state-of-the-art VPR techniques. The baselines include CNN-based methods such as NetVLAD \cite{12}, CosPlace \cite{23}, MixVPR \cite{24}, EigenPlaces \cite{24}, and BoQ \cite{30}; ViT-based methods such as SelaVPR [15], and CricaVPR [28]; and the hybrid architecture TransVPR [14]. Notably, SelaVPR and TransVPR adopt a two-stage design involving local feature-based re-ranking.
\begin{table}[htbp]
\centering
\caption{Comparison to state-of-the-art methods on benchmark datasets. The best is highlighted in bold, and the second is underlined.}
\begin{tabular}{cccccccccc}
\toprule %
\multirow{2}{*}{Methods} & \multirow{2}{*}{Backbone}                                    & \multicolumn{2}{c}{Pitts30k} & \multicolumn{2}{c}{SPED} & \multicolumn{2}{c}{Pitts250k} & \multicolumn{2}{c}{MSLS\_val} \\ \cmidrule(lr){3-4} \cmidrule(lr){5-6} \cmidrule(lr){7-8} \cmidrule(lr){9-10} 
                         &                                                              & R\@1           & R\@5          & R\@1         & R\@5        & R\@1           & R\@5           & R\@1           & R\@5           \\ \hline
NetVLAD                  & VGG16                                                        & 81.9          & 91.2         & 70.2        & 84.5       & 90.5          & 96.2          & 53.1          & 66.5          \\
CosPlace                 & VGG16                                                        & 88.4          & 94.5         & 75.5        & 87.0       & 92.3          & 97.4          & 82.8          & 89.7          \\
EigenPlaces              & VGG16                                                        & 92.5          & 96.8         & 70.2        & 83.5       & 94.1          & 97.9          & 89.1          & 93.8          \\ \hline
MixVPR                   & ResNet50                                                     & 91.5          & 95.5         & 84.7        & 92.3       & 94.2          & 98.2          & 88.0          & 92.7          \\
BoQ                      & ResNet50                                                     & 92.4          & 96.6         & 86.2        & 94.4       & 95.0          & 98.3          & \textbf{91.4}          & 94.5          \\ \hline
SelaVPR                  & ViT-L                                                        & 92.8          & 96.8         & 89.5        & 94.6       & \underline{95.2}          & \underline{98.4}          & 90.8          & \underline{96.4}          \\
CricaVPR                 & ViT-B                                                        & \underline{94.9}          & \underline{97.3}         & \underline{91.4}        & \underline{95.2}       & 95.0          & 98.1          & 90.0          & 95.4          \\ \hline
TransVPR                 & \makecell{
CNN\\   + Transformer} & 89.0          & 94.9         & 85.7        & 90.9       & 93.0          & 97.5          & 86.8          & 91.2          \\ \hline
Ours                     & \makecell{ResNet50\\   + ViT-B}  & \textbf{95.0}          & \textbf{97.8}         & \textbf{93.3}        & \textbf{97.7}       & \textbf{95.5}          & \textbf{98.6}          & \underline{91.0}          & \textbf{96.7}          \\ 
\bottomrule %
\end{tabular}
\label{tab2}
\end{table}

The quantitative results are summarized in Table \ref{tab2}. On the Pitts30k and Pitts250k datasets, our method achieves Recall@1 scores of 95.0\% and 95.5\%, respectively, significantly outperforming existing approaches and demonstrating strong robustness under large viewpoint variations. On the SPED dataset, which is characterized by considerable seasonal and lighting variations, our method also leads the field with Recall@1 and Recall@5 scores of 93.3\% and 97.7\%, respectively. These results highlight the superior environmental adaptability of the proposed local-global feature complementation network. On the more challenging cross-seasonal dataset MSLS\_val, while BoQ slightly outperforms in terms of Recall@1 (91.4\%), our method achieves a higher Recall@5 (96.7\%) and demonstrates superior overall performance over ViT-based fine-tuned methods such as SelaVPR and CricaVPR. This performance gain is primarily attributed to the introduction of frequency-domain modulation in our adapter design, which enhances the discriminative power and generalization ability of the global image representation.
\begin{table}[htbp]
\centering
\caption{Recall@1 performance on additional challenging datasets. The best is highlighted in bold and the second is underlined.}
\begin{tabular}{cccc}
\toprule
Method   & Nordland & Eynsham & St-Lucia \\ \hline
BoQ      & 85.5     & \textbf{91.5}    & \textbf{99.9}     \\
SelaVPR  & \underline{87.3}     & 89.6    & \underline{99.8}     \\
TransVPR & 63.5     & 88.5    & 98.0     \\
Ours     & \textbf{89.8}     & \underline{90.7}    & \textbf{99.9}     \\ \bottomrule
\end{tabular}

\label{tab3}
\end{table}
Additional results on other datasets are presented in Table \ref{tab3}. Our method achieves state-of-the-art Recall@1 scores of 89.8\% on Nordland, which features strong seasonal appearance changes, and 99.9\% on the urban dynamic St-Lucia dataset. On the multi-view Eynsham dataset, our method achieves 90.7\% Recall@1, slightly below BoQ but still among the best-performing methods, further validating its strong generalization under varying viewpoints.

\begin{figure}[htbp]
\centering
\includegraphics[width=\textwidth,keepaspectratio]{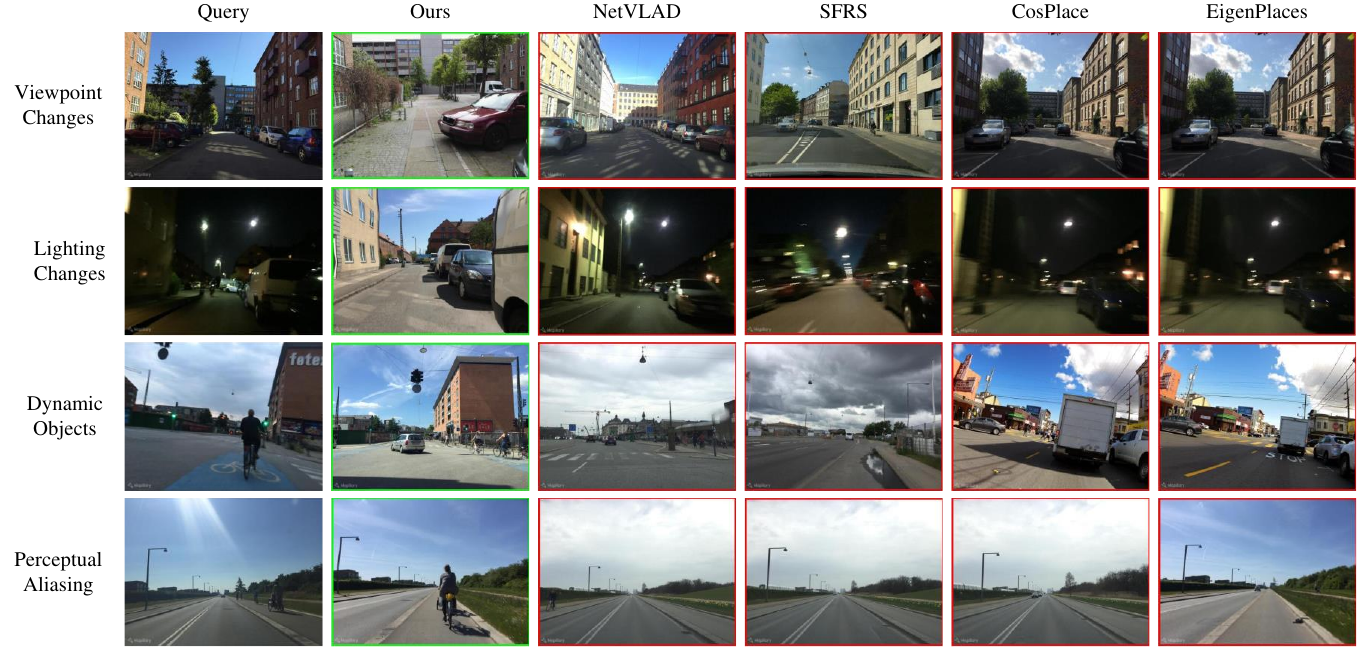}
\caption{Qualitative results. In challenging examples (perspective changes, lighting variations, dynamic objects, perceptual confusion, etc.), our method successfully returned the correct database image, while other methods all produced incorrect results.}\label{fig6}
\end{figure}

Fig. \ref{fig6} qualitatively illustrates the robustness of our method in challenging scenarios such as viewpoint shifts, lighting transitions, dynamic objects, and perceptual aliasing (similar-looking locations from different places). While existing methods struggle in these conditions, our approach consistently produces correct matches.

\begin{figure}[htbp]
\centering
\includegraphics[scale=1,keepaspectratio]{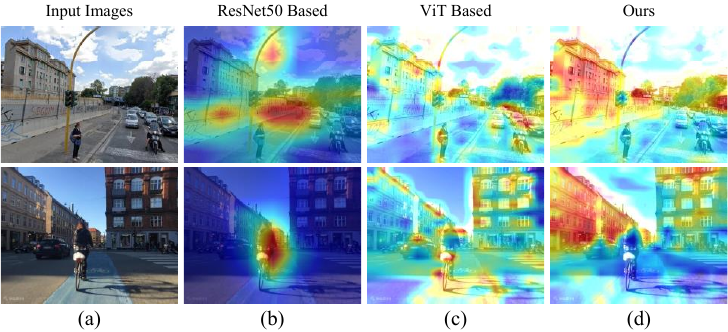}
\caption{Feature responses in urban scenes. LGCN focuses on stable structures and reduces dynamic noise, showing stronger robustness than ResNet and ViT.}\label{fig7}
\end{figure}

To further assess the discriminative capacity of our model in complex urban scenes, we visualize feature response heatmaps for different models, as shown in Fig. \ref{fig7}. The ResNet backbone tends to focus on dynamic objects such as pedestrians and cyclists. While it captures fine-grained details, it is susceptible to distraction in dynamic environments. The ViT model, on the other hand, exhibits broad response regions, reflecting its global modeling capacity, but often lacks focus on salient areas. In contrast, our LGCN model generates more concentrated and pronounced responses in structurally static regions such as building contours and lamp posts, while significantly suppressing responses to moving distractors. This advantage stems from our local-global feature complementation network and the integration of frequency-spatial modulation, which together enhance the model’s robustness and discriminative power in real-world complex environments.
\subsection{Ablation Study}
\label{subsec4_4}
To systematically validate the effectiveness of each component in the proposed method, we conduct comprehensive ablation studies on three datasets with significant environmental variations: Pittsburgh-30k, SPED, and MSLS\_val. The experiments focus on assessing the synergy among the dual-stream architecture, the dynamic feature fusion module (DFM), and the frequency-spatial adapter (FSA). All experiments share the same training settings, and performance is evaluated using Recall@1 and Recall@5, as summarized in Table \ref{tab4}. The baseline is defined as the simplest model using only the frozen ViT backbone.

To investigate the contribution of the heterogeneous dual-stream design, we implement the following variants:
\begin{itemize}
    \item +FSA: Incorporates the frequency-spatial modulation module to enhance the representational capacity of ViT features.
    \item +CNN Stream: Adds a CNN branch to the backbone and fuses it with the ViT output via simple concatenation, verifying the complementary value of heterogeneous features.
    \item +DFM: Introduces the DFM for CNN and ViT features.
    \item Full Model: Combines both DFM and FSA within the dual-stream framework, representing the complete LGCN architecture.
\end{itemize}
\begin{table}[htbp]
\caption{Comparison of different ablated versions.}
\begin{tabular}{ccccccc}
\toprule
\multirow{2}{*}{Ablated version} & \multicolumn{2}{c}{Pitts30k}  & \multicolumn{2}{c}{SPED}      & \multicolumn{2}{c}{MSLS\_val} \\ \cmidrule(lr){2-3} \cmidrule(lr){4-5} \cmidrule(lr){6-7}
                                 & R@1           & R@5           & R@1           & R@5           & R@1           & R@5           \\ \hline
Baseline   (ViT Frozen)          & 86.0          & 93.2          & 88.7          & 92.3          & 79.5          & 87.2          \\
+ FSA                        & 92.2          & 96.3          & 91.1          & 96.5          & 88.2          & 92.1          \\
+ CNN   Stream                   & 89.7          & 94.8          & 90.5          & 93.8          & 83.9          & 89.6          \\
+ DFM                           & 93.6          & 96.5          & 92.2          & 95.5          & 86.2          & 91.4          \\
Full Model   (Ours)              & \textbf{95.5} & \textbf{98.4} & \textbf{93.3} & \textbf{97.7} & \textbf{91.0} & \textbf{95.7} \\ \bottomrule
\end{tabular}

\label{tab4}
\end{table}
In the Baseline model, only the frozen ViT backbone is used for feature extraction, achieving Recall@1 scores of 86.0\%, 88.7\%, and 79.5\% on the Pittsburgh-30k, SPED, and MSLS\_val datasets, respectively, indicating limited overall performance. After introducing the Adapter module, the model shows significant improvements across all datasets, especially in Recall@1, reaching 92.2\% on Pitts30k and 88.2\% on MSLS\_val. This demonstrates that the frequency-spatial modulation mechanism effectively enhances the discriminative power of ViT features.

Further adding a CNN branch to the Baseline (+CNN Stream) also boosts performance, confirming the complementary nature between CNN and ViT features. Building on this, the incorporation of the DFM leads to additional gains, with Recall@1 increasing to 93.6\%, 92.2\%, and 86.2\% on the three datasets, respectively. This indicates that the dynamic fusion mechanism dynamically integrates heterogeneous feature streams, effectively mitigating feature redundancy and interference.

Finally, the Full Model (Ours), which integrates all modules, achieves the best performance across all evaluated datasets, with Recall@1 scores of 95.5\%, 93.3\%, and 91.0\%. Compared to the Baseline, these correspond to improvements of 9.5\%, 4.6\%, and 11.5\%, respectively, validating that the proposed dual-stream architecture, frequency-domain modeling, and adaptive fusion collectively provide strong generalization and complementarity under complex scenarios.
\section{Conclusion}
\label{sec5}
This paper addresses the limitations of single-architecture feature representation and the low efficiency of heterogeneous feature fusion in visual place recognition (VPR) tasks by proposing a local-global feature complementation network (LGCN). Our method leverages a dual-backbone architecture combining CNN and ViT, effectively integrating CNN’s strength in local structural perception with ViT’s advantage in global semantic modeling, thereby achieving comprehensive semantic representation in complex scenes. To further enhance network adaptability, we design a frequency-spatial fusion adapter that improves the feature modulation capability of a frozen ViT backbone without introducing significant additional parameters. Additionally, the proposed  a dynamic feature fusion module (DFM) dynamically adjusts the importance of CNN and ViT features based on scene context, strengthening responses in semantically salient regions (e.g., building structures) while suppressing interference from dynamic elements such as pedestrians and vehicles. Extensive experiments on seven public datasets covering urban, suburban, and cross-season scenarios validate the effectiveness and superiority of the proposed approach.

\section*{Acknowledgments}
The project is supported by Scientific Research Program Funded by Shaanxi Provincial Education Department (Program No. 22JK0303) and Natural Science Basic Research Plan in Shaanxi Province of China(Program No. 2022JQ-175).

\bibliography{sn-bibliography}

\end{document}